\documentclass[10pt,twocolumn,letterpaper]{article}

\usepackage[pagenumbers]{cvpr} %

\usepackage[dvipsnames]{xcolor}
\usepackage{xspace}
\usepackage{gensymb}
\usepackage{float}
\usepackage{tabularx}
\usepackage{microtype}
\usepackage[accsupp]{axessibility}

\definecolor{cvprblue}{rgb}{0.21,0.49,0.74}
\usepackage[pagebackref,breaklinks,colorlinks,citecolor=cvprblue]{hyperref}
\usepackage{tabularx}
\usepackage{makecell}
\usepackage{booktabs}
\usepackage{multirow}
\usepackage{multicol}
\usepackage{pifont}
\def\paperID{7399} %
\def\confName{CVPR}
\def\confYear{2024}
\usepackage{lipsum}

\newcommand{\fullname}{Decorating Untextured Shapes with Distilled Semantic Features\xspace}
\newcommand{\name}{\textsc{Diffusion 3D Features}\xspace}
\newcommand{\sname}{\textsc{Diff3F}\xspace}
\newcommand{\xmark}{\ding{55}}%

\title{\name~(\sname) \\ \fullname}

\author{%
Niladri Shekhar Dutt\textsuperscript{1,2} \quad
Sanjeev Muralikrishnan\textsuperscript{1} \quad
Niloy J. Mitra\textsuperscript{1,3} \\[1.0ex]
\textsuperscript{1}University College London \quad
\textsuperscript{2}Ready Player Me \quad
\textsuperscript{3}Adobe Research 
\\ [1.0ex]
\href{https://diff3f.github.io/}{https://diff3f.github.io/}
}

\begin{document}
\twocolumn[{%
\renewcommand\twocolumn[1][]{#1}%
\vspace*{-0.2in}
\maketitle
\vspace*{-0.2in}
\begin{center}
    \centering
    \includegraphics[width=\textwidth]{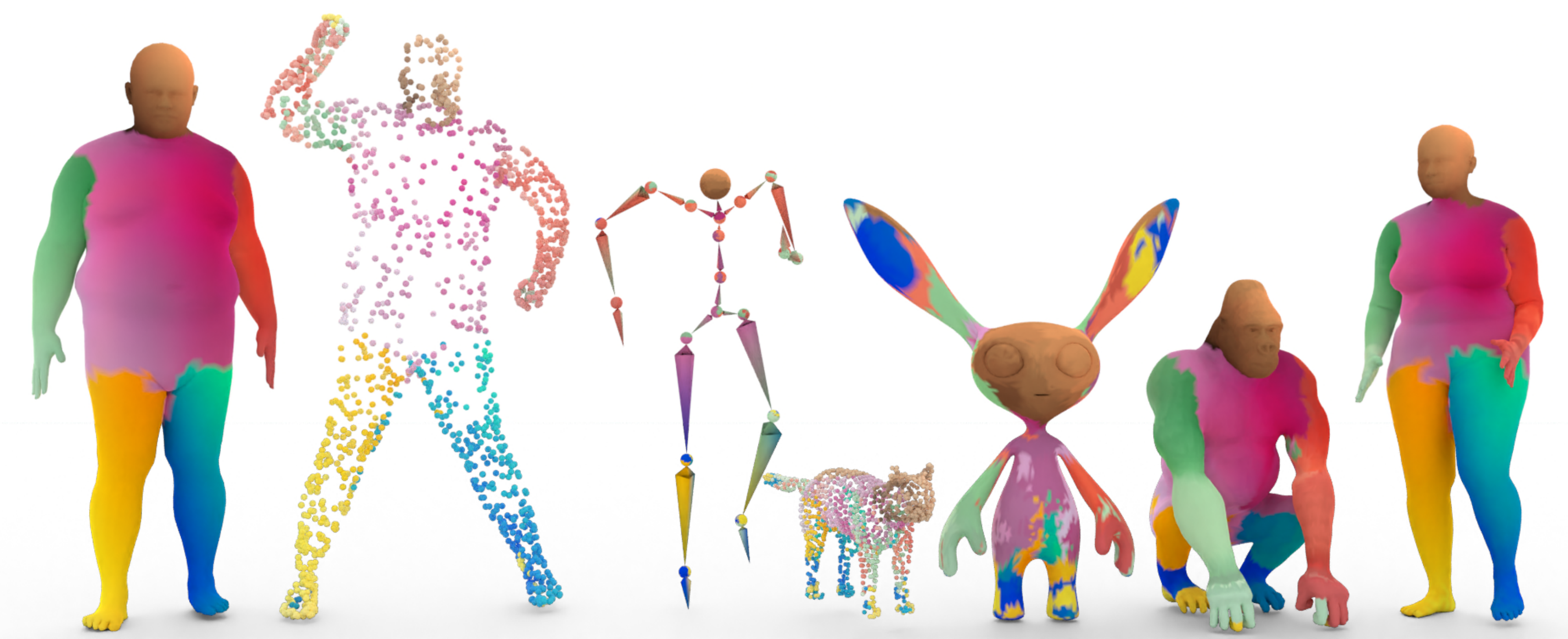}
    \captionof{figure}{\textbf{Correspondence in-the-wild.} We introduce \sname, a novel feature distiller that harnesses the expressive power of in-painting diffusion features and distills them to points on 3D surfaces. Here, the proposed features are employed for point-to-point shape correspondence between assets varying in shape, pose, species, and topology. We achieve this without any fine-tuning of the underlying diffusion models, and demonstrate results on untextured meshes, point clouds, and raw scans. The leftmost mesh is the source, and all the remaining 3D shapes are targets. Note that we show raw point-to-point correspondence, without any regularization or smoothing. Inputs are point clouds, non-manifold meshes, or 2-manifold meshes. Corresponding points are similarly colored across the shapes.
    }
    \label{fig:teaser}
\end{center}%
}]

\let\thefootnote\relax\footnotetext{Accepted at \href{https://cvpr.thecvf.com/}{CVPR'24}}

\begin{abstract}

We present \sname as a simple, robust, and class-agnostic feature descriptor that can be computed for untextured input shapes (meshes or point clouds). 
Our method distills diffusion features from image foundational models onto input shapes. 
Specifically, we use the input shapes to produce depth and normal maps as guidance for conditional image synthesis. In the process, we produce (diffusion) features in 2D that we subsequently lift and aggregate on the original surface.  
Our key observation is that even if the conditional image generations obtained from multi-view rendering of the input shapes are inconsistent, the associated image features are robust and, hence, can be directly aggregated across views. This produces semantic features on the input shapes, without requiring additional data or training. 
We perform extensive experiments on multiple benchmarks (SHREC'19, SHREC'20, FAUST, and TOSCA) and demonstrate that our features, being semantic instead of geometric, produce reliable correspondence across both isometric and non-isometrically related shape families. Code is available at \href{https://github.com/niladridutt/Diffusion-3D-Features}{https://github.com/niladridutt/Diffusion-3D-Features}.

\end{abstract}
    
\section{Introduction}
\label{sec:intro}

Feature descriptors are crucial building blocks in most shape analysis tasks. The ability to extract reliable features from input meshes or point clouds paves the way for establishing shape correspondence, extracting low-dimensional shape spaces, and learning 3D generative models, to name a few applications. Classical geometry processing algorithms, extensively explored over recent decades \cite{functionalmaps, fmnet, tombari2010unique, tosca, eisenberger2019divergence, dgcnn}, concentrate on identifying geometric invariants and, at best, coincidentally align with semantic features. Recent learning-based approaches~\cite{dpc,se-ornet,cao2023self}, trained on a limited amount of data, learn the correlation between geometric and semantic features but struggle to generalize to unseen categories. In contrast, in image analysis, a recent winner has emerged --- foundational models~\cite{clip, dino,dinov2,stable-diffusion} trained on massive image datasets have been repurposed to yield general-purpose feature descriptors~\cite{amir2021deep, dift, zhang2023tale, hyperfeatures}. Remarkably, such large-scale models have implicitly learned robust semantic features that match and often surpass classical image feature descriptors. For instance, DINO features~\cite{dinov2} and diffusion features~\cite{dift} extract dense semantic image features from photo-realistic images \textit{without} additional training. In this paper, we investigate the adaptation of this success to the realm of 3D shapes. 

A significant challenge is to address the absence of textures on most 3D models. This hinders immediate rendering to produce photo-realistic images required by image-based feature detectors mentioned earlier. Additionally, when shapes are represented as meshes, they may have non-manifold faces, making it challenging to extract UV parameterizations; when shapes are represented as point clouds, they lack connectivity information, making rendering ambiguous. One potential solution for input meshes involves utilizing recent approaches~\cite{richardson2023texture, chen2023text2tex} to first generate seamlessly textured meshes through image-guided techniques and subsequently extract image feature descriptors. For point cloud representation, one can first produce surface reconstructions \cite{screened-poisson}, then employ the aforementioned mesh-based approach. However, these methods are cumbersome, optimization-based, and unsuitable for seamless end-to-end workflows. We propose a simple and robust solution. 

We present \name~(\sname), a simple and practical framework for extracting semantic features that eliminates the need for additional training or optimization. \sname renders input shapes from a sampling of camera views to produce respective depth/normal maps. These maps are used as guidance in ControlNet~\cite{controlnet} to condition Stable Diffusion~\cite{stable-diffusion} to produce photo-realistic images. We directly use the features on these images produced during diffusion and aggregate them back on the input surfaces. Our main insight is that we do {\em not} require consistent mesh texturing to produce reliable shape features as we `denoise' smaller inconsistencies in the feature aggregation step. Since we use diffusion features, our approach reuses the intermediate information generated in the depth-guided image generation step, thus avoiding any additional training. The extracted features produce accurate semantic correspondence across diverse input shapes (see Figure~\ref{fig:teaser}), even under significant shape variations. 

We evaluate our algorithm on a range of input shapes (meshes and point clouds) and compare the quality of the extracted features on a set of established correspondence benchmarks. We study the importance of feature aggregation versus consistency of multi-view image generations, choice of the (image) features used, and robustness to input specification. We report comparable performance to state-of-the-art algorithms on multiple benchmarks, containing isometric and non-isometric variations, and outperform existing approaches in generalizability. Our main contribution is a simple and surprisingly effective semantic feature detection algorithm on 3D shapes that can be readily integrated into existing shape analysis workflows without requiring extra data or training. \sname, being semantic, is complementary to existing geometric features.

\section{Related Work}
\label{sec:related}

\begin{table}[t!]
\centering
\caption{
\textbf{Comparison of \name to related methods.}
Unlike traditional geometric feature detectors (e.g., WKS), modern learning-based approaches require training and can struggle to generalize to novel settings. We leverage strong image priors in the form of image diffusion models to directly decorate input shapes with distilled semantic features. 
}
\footnotesize %
\renewcommand{\arraystretch}{1.1}
\setlength{\tabcolsep}{3pt}
\begin{tabularx}{\linewidth}{rccccc} %
\toprule
     & \makecell{3D-CODED\\\cite{3d-coded}} & \makecell{DPC\\\cite{dpc}} & \makecell{SE-ORNet\\\cite{se-ornet}} & \makecell{FM+WKS\\\cite{functionalmaps}} & \makecell{Ours} \\
     \midrule
    No 3D training data? & \xmark & \xmark & \xmark & \checkmark & \checkmark \\
     Unsupervised? & \xmark & \checkmark & \checkmark & \checkmark & \checkmark \\
     Class agnostic? & \xmark & \xmark & \xmark & \checkmark & \checkmark \\
      Handles meshes? & \checkmark & \checkmark & \checkmark & \checkmark & \checkmark  \\
     Handles point cloud? & \checkmark & \checkmark & \checkmark & \xmark & \checkmark  \\
     \bottomrule
\end{tabularx}
\label{tab:methodVariations}
\end{table}

\paragraph{Point-to-point based shape correspondence.}
These methods, either by formulation or by explicit supervision, train algorithms to map points to points between surfaces. In other words, they establish a discrete point-to-point map instead of a continuous surface map.
3D-Coded~\cite{3d-coded} finds the correspondence between shape pairs by estimating a transformation between two point clouds. This transformation is learned by deforming a template shape to learn its reconstruction on different shapes. Elementary~\cite{elementary} extends this concept further by trying to find an ideal set of primitives to represent a shape collection. 
Many such algorithms require ground truth for training such as DCP~\cite{dcp}, RPMNet~\cite{rpmnet}, GeomFMap~\cite{geomfmap}, 3D-Coded~\cite{3d-coded}, Elementary~\cite{elementary}, and/or mesh connectivity such as GeomFMap~\cite{geomfmap} and SURFMNet~\cite{surfmnet}, both of which are hard to acquire in the real world.  

Recent efforts have focused on unsupervised methods for learning on point clouds. CorrNet3D~\cite{corrnet3d} first learns the feature embeddings using a shared DGCNN~\cite{dgcnn} and then utilizes a symmetry deformation module to learn the reconstruction and compute correspondence. DGCNN~\cite{dgcnn} uses a graph with multiple layers of EdgeConv~\cite{dgcnn} to learn feature embeddings by incorporating information from local neighborhoods. DPC~\cite{dpc} employs self and cross-reconstruction modules to learn discriminative and smooth representations and uses DGCNN for learning the per-point feature embeddings. To improve predictions for symmetrical parts, SE-ORNet \cite{se-ornet}  first aligns the source and target point clouds with an orientation estimation module before using a teacher-student model and a DGCNN backbone to find the correspondence. These methods operate directly at the geometry level and fail to understand semantic features that may not be represented directly as geometric features.

\paragraph{Surface maps based shape correspondence.}
Surface map methods learn a continuous map between two arbitrary 2-manifold surfaces. The learned map can then be sampled for point-to-point correspondence if needed. Classically, these works map eigenfunctions defined on surfaces leading to a functional mapping \cite{functionalmaps,geomfmap} or they construct an atlas of maps (charts) from $\mathbb{R}^2$ $\mapsto$ $\mathbb{R}^n$ ($n=2,3$) \cite{morreale2021neural,aigerman2022neural,mandad2017variance}. Usually, algorithms compute a specific type of surface map aiming to preserve specific geometric properties - preservation of angles \cite{levy2023least,lipman2012bounded} for conformal maps, preservation of geodesic distances \cite{sorkine2007rigid,rabinovich2017scalable} for isometric maps, etc. The unifying idea is to map both surfaces to a base domain, which can be a mesh~\cite{schreiner2004inter,kraevoy2004cross,lee1999multiresolution} or a planar region~\cite{aigerman2014lifted,aigerman2015seamless} thus mapping via the shared domain. For example, SURFMNet \cite{surfmnet} extends FMNet \cite{fmnet} to an unsupervised setting by enforcing pre-desired structural properties on estimated functional maps. These methods require meshes, often 2-manifold; our work by design can find correspondences between poorly reconstructed meshes with artifacts or directly between point clouds. Functional maps rely on geometric descriptors such as WKS~\cite{wks} to compute the mapping.

\paragraph{Multi-view rendering based learning.} Projective analysis~\cite{projectiveAnalysis:13}
encodes shapes as collections of 2D projections, performs image space analysis, and projects the results back to the 3D. This is a powerful idea. 
This family of multi-view rendering-based methods has shown remarkable performance on a variety of 3D tasks including shape/object recognition \cite{mvcnn,yu2020latent,9442303,yu2020latent,wang2022multi,tran2022self}, human pose estimation \cite{Lin_2021_CVPR}, part correspondence \cite{10.1145/3137609}, reconstruction\cite{queau2017dense}, segmentation \cite{sharma2022mvdecor,tran2022self,komarichev2019cnn,projectiveAnalysis:13} and many more. The approach involves rendering a 3D shape from multiple views and extracting information by employing visual descriptors per view, usually obtained by training a CNN in a supervised setting. Various approaches have been suggested to aggregate the features from different views like averaging, max pooling \cite{10.1145/3137609,tran2022self}, concatenating the image features, using another CNN to fuse the intermediate latent representations pooled from different views \cite{mvcnn}, etc. 

We are inspired by Huang et al.~\cite{10.1145/3137609}, who learn to aggregate descriptors by fine-tuning AlexNet \cite{alexnet} on multi-scale renders of shapes from different viewing directions. The entire network is trained in a supervised setting using contrastive loss \cite{hadsell2006dimensionality} to group semantically and geometrically similar points close in the descriptor space. The method, however, suffers from limited correspondence accuracy for lower tolerance levels (due to its noisy dataset) compared to state-of-the-art geometric methods available today.

A more recent work~\cite{abdelreheem2023zero} explored the usage of foundational image models for generating zero-shot correspondence. They use LLMs and vision models to first generate a set of segmentation maps for each object and a semantic mapping between each set. This is followed by a 3D semantic segmentation model based on SAM~\cite{kirillov2023segment} to segment the shape according to the generated set. Based on the segmented areas, geometric descriptors are initialized to compute a functional map. Finally, they apply iterative refinement to produce a final point-to-point correspondence. Instead, we distill an image foundational model to produce descriptors with rich semantic features that can be directly used to compute correspondence.

\vspace{-3mm}
\paragraph{Aggregating 3D features from 2D foundational models.} 3D Highlighter~\cite{decatur20233d} renders a mesh from multiple views and calculates their CLIP~\cite{clip} embedding. %
Distilled Feature Field~\cite{kobayashi2022decomposing} distills CLIP or DINO embeddings to 3D feature fields to enable zero-shot segmentation of Neural Radiance Fields. NeRF Analogies \cite{fischer2024nerf} further shows that DINO features on multi-view rendered images can be used to calculate correspondence for semantically transferring the appearance of a source NeRF to a target 3D geometry. We utilize ControlNet~\cite{controlnet} in-the-loop to generate multi-view inconsistent textured renderings and, in that process, generate diffusion features that are semantically coherent to compute accurate correspondence.

\begin{figure*}[ht!]
    \centering
    \includegraphics[width=\textwidth]{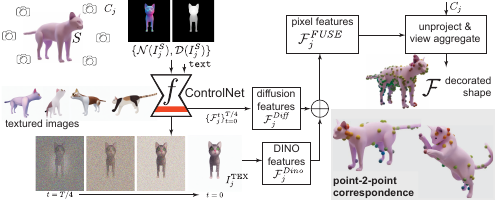}
    \caption{\textbf{Method overview.} \sname is a feature distiller to map semantic diffusion features to 3D surface points. We render the given shape without textures from multiple views, and the resulting renderings are in-painted by guiding ControlNet with geometric conditions; the generative features from ControlNet are fused with features obtained from the textured rendering, followed by unprojecting to the 3D surface. Note that the textured images obtained by conditioning ControlNet from different views can be inconsistent. }
    \label{fig:overview}
    \vspace{-\baselineskip}
\end{figure*}

\section{Method} \label{sec:method}
We aim to decorate 3D points of a given shape in any modality -- point clouds or meshes -- with rich semantic descriptors. Given the scarcity of 3D geometry data from which to learn these meaningful descriptors, we leverage foundational vision models trained on very large datasets to obtain these features. This enables \sname to produce semantic descriptors in a zero-shot way. Our code can be found \href{https://github.com/niladridutt/Diffusion-3D-Features}{here}.

\subsection{Semantic Diffusion Features}  \label{sec:distill}
Given a shape $S$ with vertices $V \in \mathbb{R}^3$, we want to project it to the image space to distill per-point semantic 3D features from images. We define an image projector $P$ as
\begin{equation} \label{eq:rendereq}
    P(\cdot | C_j) := S \mapsto I^S_j \in \mathbb{R}^{H \times W}, %
\end{equation}
where $H, W$ denote the height and width of the image rendered by 
$P$, with $C_j$ representing the $j^{th}$ camera producing the image $I^S_j$. $P$ can be a renderer with shading or simply a rasterizer that returns the depth from the camera.

As an emergent behaviour, pre-trained foundational vision models have been found to assign distinctive semantic features~\cite{dift} to pixels in the input image, to be able to distinguish between nearby pixels to perform core tasks like object detection or segmentation. Our core idea to get such features is, therefore, to drive a pretrained foundational model to perform a challenging task that requires generating semantic per-pixel features during the process, so that we can extract these features into 3D. Since we aim for per-point features, instead of regional descriptors, we add textures to rendered images conditioned on text prompts. Creating a realistic textured image from an untextured image requires the model to distinguish between nearby pixels in their semantic sense such that the visual result satisfies the text prompt. For example, a shading model may color a drawn cube completely gray. Still, when conditioned on the text ``iron box", it would be driven to add specific characteristics, such as metallic textures that clearly allow it to be identified as ``iron". 

Given a point cloud or raw mesh, projection $P$ produces an untextured image with silhouette or shading, respectively. We drive a diffusion model~\cite{stable-diffusion} $f$ to color textureless $I_j^S$ (from camera view $C_j$) realistically and take it to the RGB space:
\begin{equation}
    f(\cdot | G,\texttt{text}) := I^S_j \in \mathbb{R}^{H \times W} \mapsto I_{j}^{\textit{TEX}} \in \mathbb{R}^{H \times W \times 3}, 
\end{equation}
where $G$ is a set of functions describing geometric constraints that guide the texturing model and $\texttt{text}$ defines the text prompt defining the subjects. We guide the texturing by providing constraints $G$ to ControlNet~\cite{controlnet}. In effect, $f$ projects shape $S$ to an RGB image based on camera $C_j$. %

\begin{figure*}[t!]
    \centering
    \includegraphics[width=\linewidth]{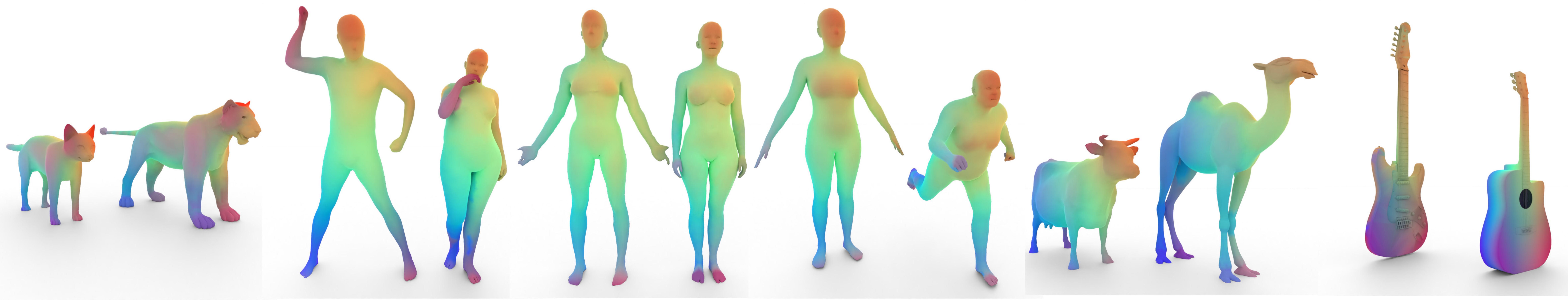}
    \caption{\textbf{Results gallery.} \sname's performance on various point correspondence challenges. Corresponding points are similarly colored. Note that \sname can successfully distinguish between symmetric parts and remains fairly robust under pose and shape variations. For each shape pair, the source is on the left and the target is on the right.}
    \label{fig:results}
\end{figure*}

\subsection{Semantics through Painting} \label{sec:painting}
Realistically texturing a silhouette image is an open problem and a challenging task. Given untextured images $\{I^S_j\}$, a naive approach of assigning a constant color does not require inferring the semantics of the given geometry. We, therefore, condition our painting module $f$ with geometric constraints that describe the latent 3D object.

We define $G$ as a set of geometric maps that can be applied as conditional image constraints,
\begin{equation} \label{eq:coloreq}
    G := \{\mathcal{N}(I^S_j),\mathcal{D}(I^S_j)\},
\end{equation}
where $\mathcal{N}$ is a normal map and $\mathcal{D}$ is a continuous depth map from the camera describing the input shape. When combined with a text prompt, we expand \Cref{eq:coloreq} as
\begin{equation} \label{eq:fullcoloreq}
    f(\cdot | \mathcal{N}(I^S_j),\mathcal{D}(I^S_j),\texttt{text}) := I^S_j \mapsto I_{j}^{\textit{TEX}}.
\end{equation}

During this texturing forward pass, we extract features $\mathcal{F}^{t}_{L}$ from an intermediate layer $L$ of Stable Diffusion's UNet decoder at diffusion time step $t$ with $t\in [0,T]$. We use DDIM~\cite{ddim} to accelerate the sampling process for Stable Diffusion~\cite{stable-diffusion} and use 30 inference steps. For notational simplicity, in the following, we drop the layer index $L$, i.e., we use $\mathcal{F}^{t}_{j}$ to indicate $\mathcal{F}^{t}_{L,j}$. We thus directly get our feature renderer as, 
 \begin{equation} \label{eq:features}
    f(\cdot | \mathcal{N}(I^S_j),\mathcal{D}(I^S_j),\texttt{text},L,t) := I^S_j \mapsto \mathcal{F}^{t}_{j}, %
\end{equation}
These features are normalized to have unit norm:
\begin{equation} \label{eq:norm}
    \mathcal{F}^{t}_{j} := {\mathcal{F}^{t}_{j}}/{\|\mathcal{F}^{t}_j\|_2}.
\end{equation}

 We aggregate the normalized features using a weighted approach starting from T/4 to the final denoising step $0$. The weights $w_t$ are linearly spaced from 0.1 to 1 depending on the number of time steps ($t$) to assign higher weights to the embeddings from less noisy images (i.e., higher $t$).
Each pixel gets a $1280$ dimensional feature from the diffusion UNet, aggregated over diffusion time steps. Specifically, 
\begin{equation}
    \mathcal{F}^\textit{Diff}_j := \sum_{t=0}^{{T}/{4}} w_t \cdot \mathcal{F}^{t}_{j} \in \mathbb{R}^{H \times W \times 1280}.
\end{equation}
 We further fuse the diffusion features $\mathcal{F}_j$ with DINOv2~\cite{dinov2} features $\mathcal{F}^\textit{Dino}_j$ extracted from the textured renderings $I_{j}^{\textit{TEX}}$. We also normalize these image features as in Equation~~\ref{eq:norm}. It has been noted that DINO features contain strong complementary semantic signals but are weaker regarding spatial understanding \cite{zhang2023tale}. Hence, combining the two features gives stronger semantic descriptors while retaining spatial information. We employ a  feature fusion strategy proposed by \cite{zhang2023tale}, where we first normalize the features and then concatenate them as,
\begin{equation}
    \mathcal{F}^\textit{FUSE}_{j} := (\alpha \mathcal{F}^\textit{Diff}_j, (1 - \alpha)\mathcal{F}^\textit{Dino}_j)
\end{equation}
where $\alpha$ is a tunable parameter; we use $\alpha=0.5$ in all our experiments. $\mathcal{F}^\textit{FUSE}_{j}$ is also unit-normalized as in Equation~\ref{eq:norm}.

\subsection{Distilling 2D Features to 3D}  \label{sec:unproject}
We leverage known camera parameters to unproject features from the image space back to the points on the 3D input, i.e.,  \(\mathcal{F}^{\textit{FUSE}}_j \xrightarrow{P^{-1}} \mathcal{F}^{\textit{3D}}_j\), where $P$ is the projection function from equation \ref{eq:rendereq}. We also employ a ball query $B_r(x)$, introduced in \cite{pointnet++}, to facilitate feature sharing within local neighborhoods of radius $r$ around any surface point $x \in S$ and promote local consensus. We use $r=1\%$ of the object's bounding box diagonal length. This is particularly useful for shape correspondence where points in a local neighborhood of the source should match with points in a local neighborhood of the target. 

To aggregate features from multiple views per point, we compute the mean of the normalized feature vectors. We also experimented with max pooling, but the results were inferior. Our rendering setup and unprojection with ball querying make the per-point accumulated features coherent, enabling a simple aggregation. Moreover, we render the 3D shape from several views ($n = 100$), which further stabilizes the aggregation, resulting in descriptors that mostly capture semantic meaning:
\begin{equation}
    \mathcal{F} := \frac{1}{n}\sum_{j=1}^{n} \mathcal{F}^{3D}_j.
\end{equation}
The above step aggregates descriptors per vertex across $n$ views, spread uniformly around a sphere around the object, to compute our final semantic 3D point descriptors. Next, we describe how to use these descriptors to compute correspondences between pairs of shapes.

\subsection{Computing Correspondence} \label{sec:correspondence}
Given a source point cloud $\mathcal{S}$ and a target point cloud $\mathcal{T}$, we want to find a mapping $m : \mathcal{S} \mapsto \mathcal{T}$, such that we compute a corresponding point $\mathcal{T}_k \in \mathcal{T}$ for each point $\mathcal{S}_i \in \mathcal{S}$ where $1 \leq i,k \leq N$.

\begin{table*}[t!]
\caption{ \textbf{Comparison.}  We report correspondence accuracy within 1\% error tolerance, with our method against competing works. The Laplace Beltrami Operator~(LBO) computation for Functional Maps is unstable on TOSCA since the inputs contain non-manifold meshes. By ‘*’ we denote results reported by SE-ORNet \cite{se-ornet}.}
    \label{tab:comparison_of_methods}
    \resizebox{\linewidth}{!}{
\begin{tabular*}{1.07\linewidth}{@{\extracolsep{\fill}}  c | c|c | c|c | c|c | c|c | c|c | c|c }
    \hline
    \textbf{Method} $\mapsto$  & \multicolumn{2}{c|}{DPC~\cite{dpc}} & \multicolumn{2}{c|}{SE-ORNet~\cite{se-ornet}} & \multicolumn{2}{c|}{3DCODED~\cite{3d-coded}} & \multicolumn{2}{c|}{FM~\cite{functionalmaps}+WKS~\cite{wks}} & \multicolumn{2}{c|}{\sname (ours)} & \multicolumn{2}{c}{\sname (ours)+FM~\cite{functionalmaps}}  \\
    $\downarrow$ \textbf{Dataset}   & $acc \uparrow$ & $err \downarrow$ & $acc \uparrow$ & $err \downarrow$ & $acc \uparrow$ & $err \downarrow$ & $acc \uparrow$ & $err \downarrow$ & $acc \uparrow$ & $err \downarrow$ & $acc \uparrow$ & $err \downarrow$ \\
    \hline  \hline
    TOSCA & \underline{\textit{30.79}} & \textbf{3.74} & \textbf{33.25 }& \underline{\textit{4.32}} & 0.5* & 19.2* & \xmark & \xmark & 20.27 & 5.69 & \xmark & \xmark \\
    SHREC'19 & 17.40 & 6.26 & 21.41 & 4.56 & 2.10 & 8.10 & 4.37 & 3.26 & \textbf{26.41} &\underline{\textit{1.69}} & \underline{\textit{21.55}} & \textbf{1.49}  \\
    SHREC'20 & 31.08 & 2.13 & 31.70 & 1.00 & \xmark & \xmark & 4.13 & 7.29 & \textbf{72.60} & \underline{\textit{0.93}} & \underline{\textit{62.34}} & \textbf{0.71} \\
    \hline
\end{tabular*}}
\end{table*}

\noindent \textit{Point-to-Point:} To find correspondence between points of two shapes, we compute their point descriptors independently and match them by cosine similarity in the shared feature space. We calculate similarity as the cosine of the angle between the source ($\mathcal{S}$) and target ($\mathcal{T}$) feature vectors, $\mathcal{\mathcal{F}_S}$ and $\mathcal{\mathcal{F}_T}$, respectively. Specifically, 
\begin{equation}
    s_{ik} := \frac{\left<\mathcal{F}_{\mathcal{S}_i}, \mathcal{F}_{\mathcal{T}_k}\right>}{\|\mathcal{F}_{\mathcal{S}_i}\|_2 \|\mathcal{F}_{\mathcal{T}_k}\|_2}
\end{equation}
where $\mathcal{F}_\mathcal{S}^i$ and $\mathcal{F}_\mathcal{T}^j$ are the $i^{th}$ and $k^{th}$ rows of $\mathcal{F}_\mathcal{S}$ and $\mathcal{F}_\mathcal{T}$ respectively, $\left<\cdot\right>$ denotes the dot product operation. We choose a corresponding point $\mathcal{T}_k$ in $\mathcal{T}$ for every point $\mathcal{S}_i$ in $\mathcal{S}$ where $s_{ik}$ is highest. Note that, in order to assess the quality of our decorated features, we assign correspondence based on the highest per-vertex similarity and do not regularize the solution with any other energy terms. 

\noindent \textit{Surface-to-Surface:} While point-to-point correspondence is important for certain applications like non-rigid registration, we note that it might also be desirable in certain cases to compute a continuous surface-to-surface map, rather than matching discrete points. To enable this, we pass our computed descriptors to a vanilla Functional Map \cite{functionalmaps} implementation, which returns a continuous surface-to-surface map that can then be used for direct correspondences.

\section{Evaluation}\label{sec:evaluation}

\begin{table}[!b]
\caption{ \textbf{Generalization.} We compare generalization capabilities of \sname vs others by training on one dataset and testing on a different set. For DPC and SE-ORNet, we choose SURREAL and SMAL as the training sets for human and animal shapes, respectively -- these larger datasets  lead to improved generalization scores. By `*' we denote results reported by SE-ORNet~\cite{se-ornet}.}
    \centering
    \resizebox{1.01\linewidth}{!}{
    \begin{tabular}{c|c|c|c|c|c|c|c}
    \hline
        \multirow{2}{*}{\textbf{Train}} & \multirow{2}{*}{\textbf{Method}} & \multicolumn{2}{c|}{TOSCA} & \multicolumn{2}{c|}{SHREC'19} & \multicolumn{2}{c}{SHREC'20} \\
        \cline{3-8}
        ~ & ~ & $acc\uparrow$ & $err\downarrow$ & $acc\uparrow$ & $err\downarrow$ & $acc\uparrow$ & $err\downarrow$ \\ \midrule  \midrule
        \multirow{2}{*}{SURREAL} & DPC~\cite{dpc} & 29.30 & \underline{\textit{5.25}} & 17.40 & 6.26 & 31.08 & 2.13 \\
        \cline{2-8}
        ~ & SE-ORNET~\cite{se-ornet} & 16.71 & 9.19 & \underline{\textit{21.41}} & \underline{\textit{4.56}} & \underline{\textit{31.70}} & \underline{\textit{1.00}}\\
         \hline
        \multirow{2}{*}{SMAL} & DPC~\cite{dpc} & \underline{\textit{30.28}} & 6.43 & 12.34 & 8.01 & 24.5* & 7.5*\\
        \cline{2-8}
        ~ & SE-ORNET~\cite{se-ornet} & \textbf{31.59} & \textbf{4.76} & 12.49 & 9.87 & 25.4* & 2.9* 
         \\ \hline
        Pretrained & \sname (ours) & 20.27 & 5.69 & \textbf{26.41} & \textbf{1.69} & \textbf{72.60} & \textbf{0.93} \\\hline
    \end{tabular}}
    \label{table:generalization}
\end{table}

\subsection{Datasets and Benchmarks}
We evaluate our method on datasets involving both human and animal subjects to showcase the efficacy and applicability of our approach. To make a fair comparison with existing works, we follow a similar experiment setup described in DPC \cite{dpc} and SE-ORNet \cite{se-ornet}. 

\textbf{Human shapes}: We test our method on SHREC'19 \cite{shrec'19} comprising of 44 actual human scans, which are organized into 430 annotated test pairs with considerable variation. We choose the more challenging remeshed version from \cite{geomfmap}. We also evaluate our method on the FAUST benchmark~\cite{faust}.

\textbf{Animal shapes}: For testing our method on animal shapes, we evaluate our method on the SHREC'20~\cite{shrec20} and TOSCA~\cite{tosca} datasets. SHREC'20 contains various animals in different poses with non-isometric correspondence annotated by experts. We compute correspondence on the annotated correspondences per pair (approximately 50). TOSCA comprises of 80 objects representing a mixture of animals and humans, formed by deforming template meshes. We ignore the human figures and select all the 41 animal figures within the test set and pair shapes from the same category to create a testing set of 286 paired samples.

\subsection{Evaluation Metrics} \label{sec:metric}
We use the average correspondence error and the correspondence accuracy as our evaluation criteria. Since our method operates in the domain of point clouds, we use a Euclidean-based measure, as used in previous works~\cite{corrnet3d, dpc, se-ornet}.

The average correspondence error for a pair of shapes, source $\mathcal{S}$ and target $\mathcal{T}$ is defined as:
\begin{equation}
    err = \frac{1}{n} \sum_{\mathcal{S}_i \in \mathcal{S}} \| f(\mathcal{S}_i) - t_{gt}\|_2
\end{equation}
where $f(\mathcal{S}_i)$ is the computed correspondence for $\mathcal{S}_i \in \mathcal{S}$ in $\mathcal{T}$ and $t_{gt} \in \mathcal{T}$ is the ground truth correspondence for a set of $n$ samples. 

The correspondence accuracy is measured as the fraction of correct correspondences within a threshold tolerance distance:
\begin{equation}
   acc(\epsilon) = \frac{1}{n} \sum_{\mathcal{S}_i \in \mathcal{S}} \mathbb{I} (\| f(\mathcal{S}_i) - t_{gt}\|_2 < \gamma d)
\end{equation}
where $\mathbb{I}(\cdot)$ is the indicator function , $\gamma \in [0,1]$ is the error tolerance, and $d$ is the maximal Euclidean distance between points in $T$.

\subsection{Baseline Methods}
We present our results on input pairs of meshes and pairs of pointclouds. We compare our method to recent state-of-the-art methods in shape correspondence namely DPC \cite{dpc}, SE-ORNet \cite{se-ornet}, and 3D-CODED~\cite{3d-coded}. While 3D-CODED requires an extensively annotated dataset for training, DPC and SE-ORNet are unsupervised methods and have been trained on human datasets- SURREAL~\cite{surreal} and SHREC'19~\cite{shrec'19}, as well as animal datasets- SMAL~\cite{smal} and TOSCA~\cite{tosca}. Note that we do not have access to pretrained 3D-CODED models for animal models.  In comparison, our method requires no training, enabling zero-shot feature extraction. Our method computes semantic descriptors, therefore, for a complete comparison, we also evaluate against the Wave Kernel Signature (WKS)~\cite{wks11} geometric descriptors combined with Functional Maps~\cite{functionalmaps}.

\subsection{Evaluation on Human Shapes}\label{sec:shrec}
We present results on the SHREC'19 dataset. \Cref{tab:comparison_of_methods} shows correspondence accuracy at 1\% error tolerance which represents a near-perfect hit. Our method achieves a state-of-the-art correspondence accuracy of 26.41\% at 1\% error tolerance, an improvement of 5\%.

\begin{figure}[b!]
    \centering
    \includegraphics[width=0.7\linewidth]{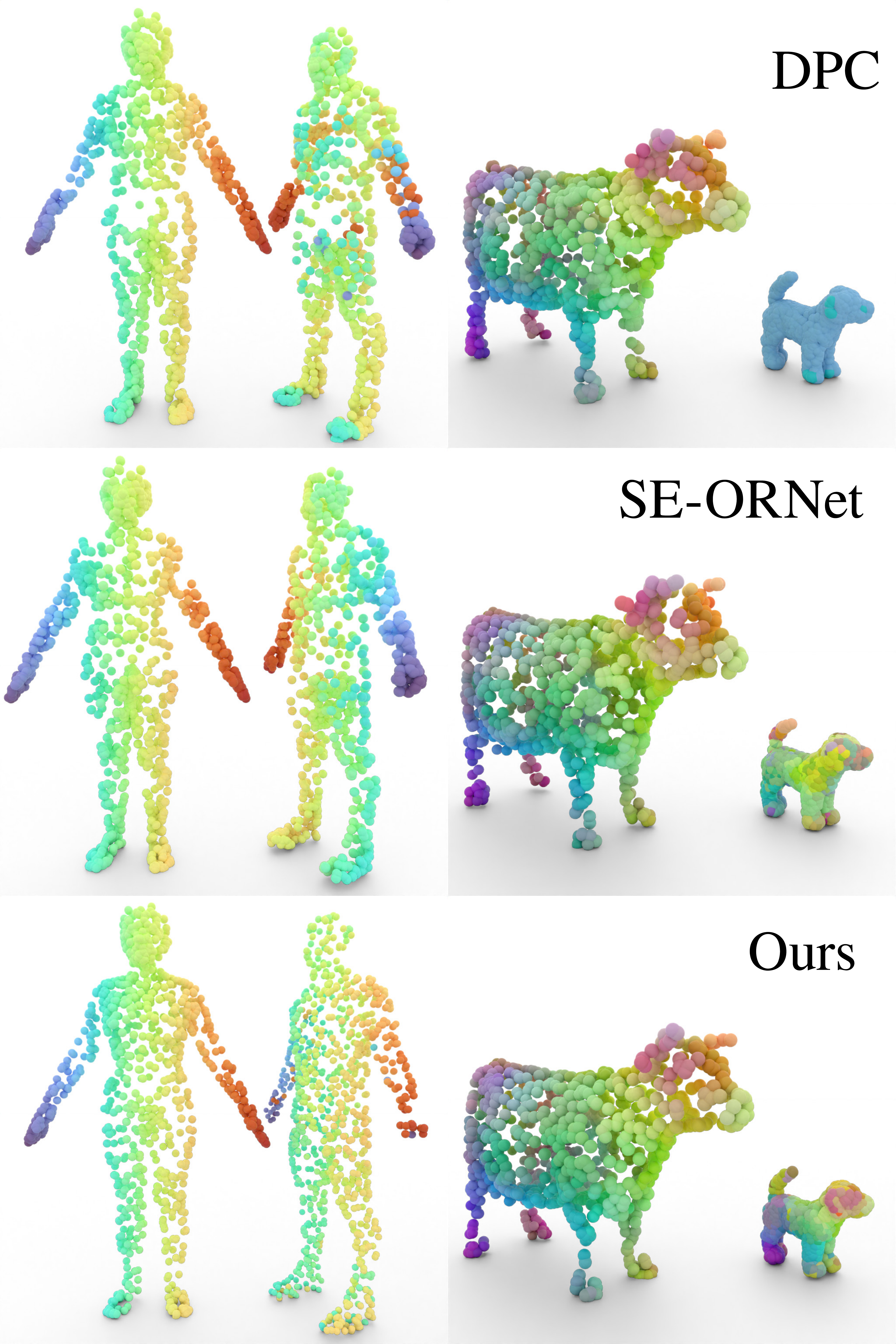}
    \caption{\textbf{Comparisons.} We compare our \sname~(bottom) against SOTA methods (i.e., DPC~\cite{dpc} and SE-ORNet~\cite{se-ornet}) for the task of point-to-point shape correspondence. Corresponding points, computed as described in Section~\ref{sec:correspondence}, are similarly colored. We show results using point cloud rendering of our method for the human pair (left) and results with mesh rendering for the animal pair (right). \Cref{tab:comparison_of_methods} shows qualitative evaluation on benchmarks.  }
    \label{fig:comparisons}
\end{figure}

Many tasks, including alignment and texture transfer, require a certain number of precise correspondences rather than average correspondence quality to work well. We choose baseline methods trained on SURREAL as it is a significantly larger dataset (consisting of human shapes) than SHREC'19, leading to improved performance. Our method achieves the highest correspondence accuracy compared to existing works and the lowest average correspondence error compared to baseline methods, as seen in \Cref{tab:comparison_of_methods}. We show qualitative results for comparison in \Cref{fig:comparisons} using our method with point cloud rendering. While DPC and SE-ORNet both get confused by the different alignments of the human pair resulting in a flipped prediction, ours, being a multi-view rendering-based method, it is robust to rotation. Hence, it can reliably solve the correspondence. We show additional visual results from a human source to multiple targets spanning modality, class and pose in \Cref{fig:teaser}. Most correspondences including highly non-isometric deformations are accurate but we see misaligned correspondence for the human to alien pair as the legs get flipped. This is because the front and back sides are less clear as the mesh has no dominant front feature. Moreover, we do not perform any processing on the raw meshes and use random coordinate system. Please refer to the supplemental for similarity heatmaps for selected points on examples. 

\noindent\textbf{Evaluation on FAUST scans.} We further evaluate \sname on the FAUST~\cite{faust} intra-subject challenge, which consists of high-resolution human scans (100k+ vertices). \sname achieves an average geodesic error of 5.29cm. Error profiles and visual results can be found on the FAUST website.

\subsection{Evaluation on Animal Shapes}
\vspace{-2mm}
We evaluate baseline methods trained on TOSCA and SMAL, and select the best performing configuration- for DPC and SE-ORNet, we choose TOSCA, whereas for 3D-CODED we choose SMAL. Our method achieves comparable accuracy and error to the baseline methods on the TOSCA dataset, as seen in \Cref{tab:comparison_of_methods}, and generalizes better than baseline methods trained on human shapes, as seen in \Cref{table:generalization}. Results using 3D-CODED are particularly poor on TOSCA mainly for two reasons: (i)~It needs a much larger dataset with ground truth annotations, which is not available for animal shapes; and (ii)~it computes correspondence through template deformation, which fails on TOSCA due to the varied shapes and poses of different animals. TOSCA consists of highly isomteric shapes, therefore we also evaluate our method on SHREC'20, which consists of highly non-isometric pairs of animals. We outperform baseline methods by a large margin for non-isometric shapes thanks to the semantic nature of \sname. While the evaluation is on a subset of only about 50 points as the number of annotated points is very limited, we show dense correspondence in \Cref{fig:comparisons}. The visual results showing dense correspondence for non-isometric pairs highlight the efficacy of our semantic descriptors compared to competing methods. DPC can get confused by the radical change in structure, while SE-ORNet has largely misaligned correspondences. We present additional visual results of animal pairs and a guitar pair in \Cref{fig:results}.

\subsection{Ablations}

We ablate different components of our method and report their performance. Table~\ref{table:ablation} shows our findings. We find that adding realistic texture, as opposed to only shading, results in a significant improvement in terms of accuracy and reducing errors. We also explore a baseline method using DINO features on consistent textures obtained using TEXTure~\cite{richardson2023texture}. Ours is better, particularly at an \textit{accuracy} of 1\% $err$ tolerance, as diffusion features capture more geometric information than DINO. Additionally, varied textured renderings enable a more robust feature aggregation due to the implicit denoising of unnecessary feature dimensions such as color. We note that TEXTure yielded poor results for humans. As meshes are not aligned and rely on iterative inpainting, if the first texture is poor, subsequent textures are poor, too. In contrast, ours aggregates over multiple views. Although our complete approach produces the second-best score in every category, incorporating all of our parts together (including fusion with DINO) resulted in the best overall balance of high accuracy and low average error.

\begin{table}[t!]
\caption{\textbf{Ablation.} We ablate different components of our method and compare accuracy at 1\% tolerance on SHREC'19 and SHREC'20, against our full method (last row).}
\centering
\resizebox{\columnwidth}{!}{
\begin{tabularx}{\linewidth} {c|c|c|c|c}
\footnotesize 
    \textbf{Ablation} & \multicolumn{2}{c|}{SHREC'19} & \multicolumn{2}{c}{SHREC'20}  \\
    & $acc \uparrow$ & $err \downarrow$ & $acc \uparrow$ & $err \downarrow$ \\
    \midrule  \midrule
    w/o ControlNet \\(untextured) & 17.20 & 2.04 & 65.48 & \textbf{0.69} \\
    \hline
    TEXTure\cite{richardson2023texture}+DINO & 17.20 & 2.04 & 65.48 & \textbf{0.69} \\
    \hline
    w/o Fusion with DINO & \textbf{26.53} & 2.06 & 64.89 & 1.60 \\
    \hline
    w/o Normal Maps & 25.68 & \textbf{1.67} & 69.71 & 1.17 \\
    \hline
    w/o Time Aggregation & 25.73 & 1.71 & 68.95 & \underline{\textit{0.87}} \\
    \hline
    w/o Ball query & 25.72 & 1.73 & \textbf{74.10} & 0.99 \\
    \hline
    \sname (full method) & \underline{\textit{26.41}} & \underline{\textit{1.69}} & \underline{\textit{72.60}} & 0.93 \\
    \bottomrule
\end{tabularx}}
    \label{table:ablation}
\end{table}

\begin{figure}[h!]
    \centering
    \includegraphics[width=.7\linewidth]{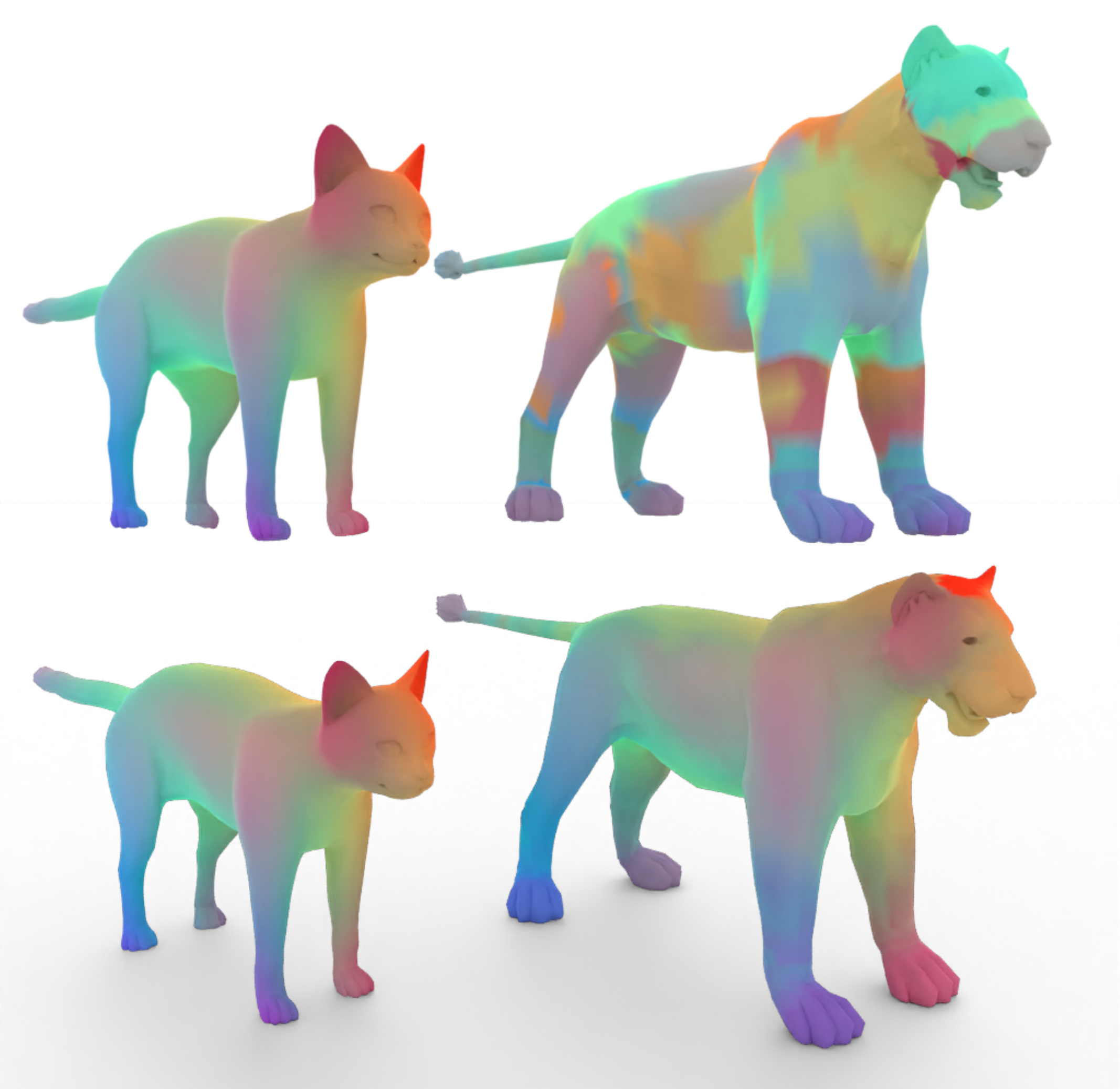}
    \caption{\textbf{Regularizing point-2-point maps.} We compare the effectiveness of vanilla functional maps with the Wave Kernel Signature as descriptors (top) vs our descriptors \sname (bottom). Ours being semantic enables Functional Maps to work with non-isometric deformations even though FMs typically struggle with such cases when using traditional geometric descriptors. Our descriptors yield accurate correspondence in most cases, thus eliminating the need for further refinement algorithms typically used in related works. }
    \label{fig:comparisons}
    \vspace{-\baselineskip}
\end{figure}

\section{Part Segmentation}
\begin{figure}[t!]
\vspace*{0.2in}
  \includegraphics[height=0.2\columnwidth]{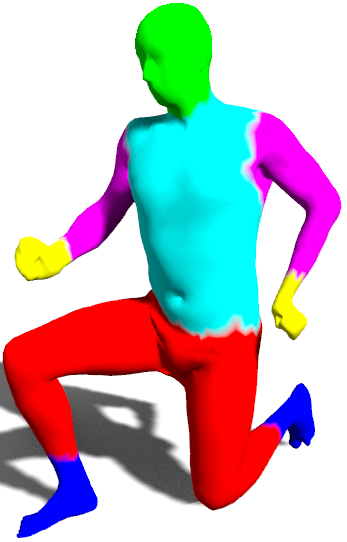}
  \includegraphics[height=0.19\columnwidth]{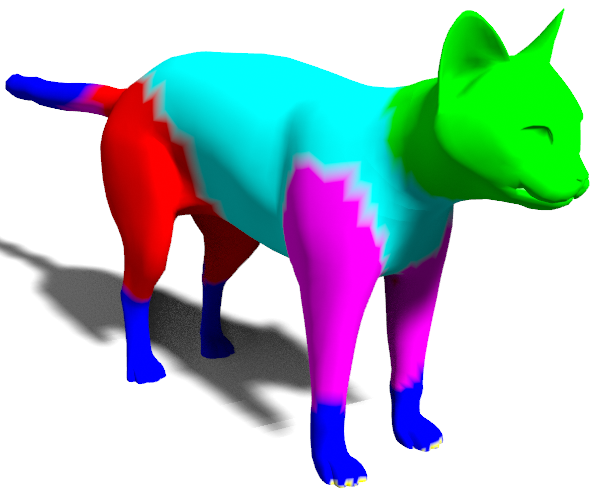}
  \includegraphics[height=0.17\columnwidth]{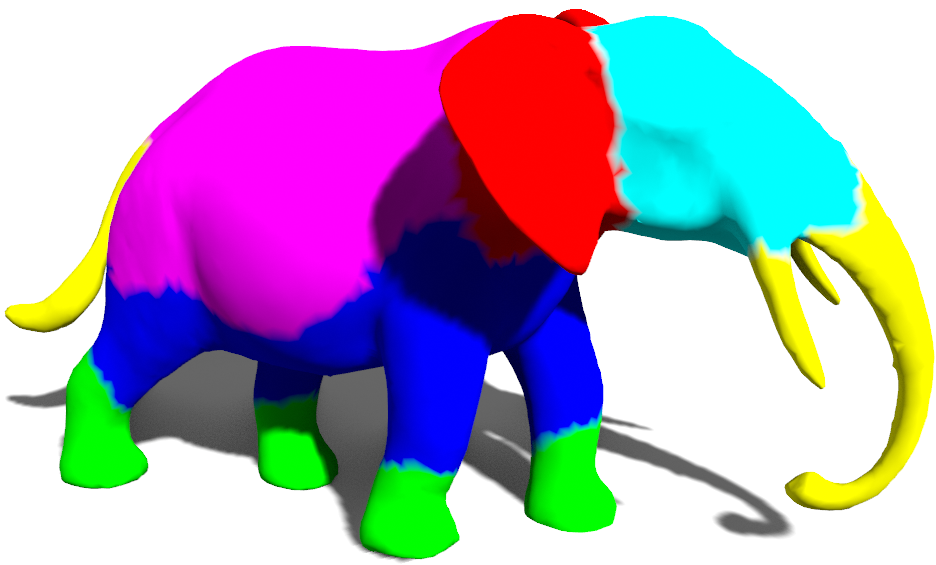}
  \includegraphics[height=0.2\columnwidth]{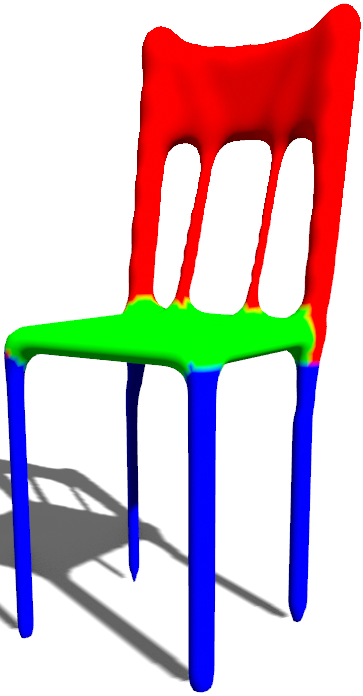}
  \includegraphics[height=0.18\columnwidth]{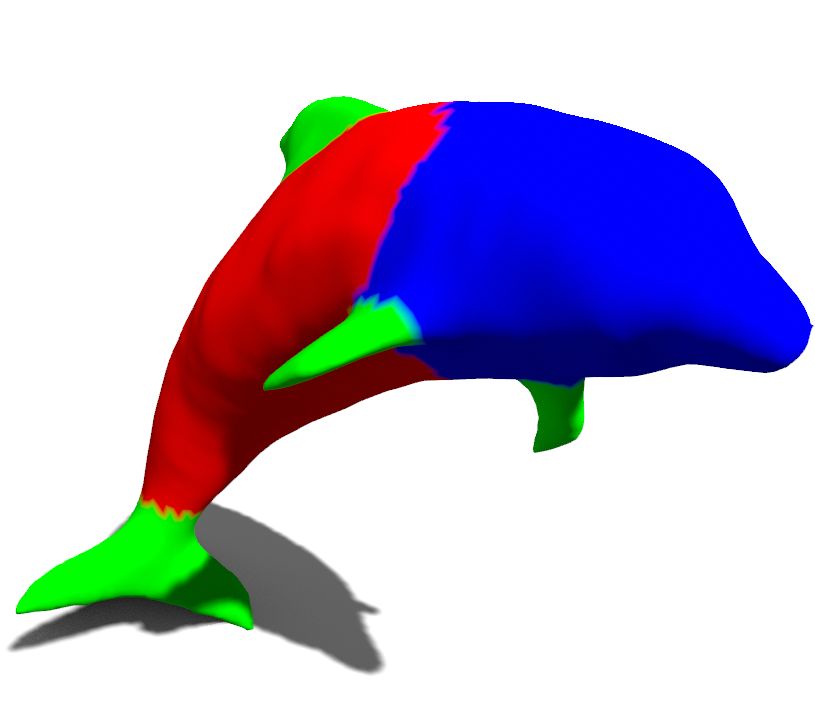}
   \caption{\textbf{Segmentation.} We apply k-means on \sname features with $k=6$ on human and elephant; and $k=3$ on chair and dolphin. The cat is segmented using k-means centroids of the human leading to corresponding part segmentation (arms map to front legs, etc.).}
   \label{fig:segment}
\end{figure}

We directly apply k-means clustering to our features to extract part segments. Interestingly, we discover that the k-means centroids, extracted from one shape (e.g., human), can be used to segment another (e.g., cat), thanks to the semantic nature of our descriptors. This leads to corresponding part segmentation (arms of the human map to the front legs of the cat, head maps to head, etc.) as seen in \Cref{fig:segment}. One possible method to automatically identify the number of segments ($k$) is to query an LLM, as explored in~\cite{abdelreheem2023zero}.

\section{Conclusion}
\vspace{-1mm}
We introduced \sname as a robust semantic descriptor for textureless input shapes like meshes or point clouds. Distilled through image diffusion models, these descriptors, without the need for extra data or training, complement existing geometric features and generalize well across diverse inputs. Our thorough evaluation on benchmark datasets, including isometric and non-isometric shapes, positions \sname as a state-of-the-art performer. We outperform recent learning-based methods on multiple datasets, demonstrating superior performance even on shapes beyond the training sets.

\vspace{-4mm}

\paragraph{Limitations.} Since our method relies on multi-view images, \sname fails to produce features on parts of the shapes that are invisible from all the sampled views (self-occlusion). Further, since we aggregate (diffusion) features from image diffusion models, we inherit their limitations in terms of suffering from bias in the dataset and/or view bias for objects. For example, the features aggregated on a horse model are worse in less seen regions, like the underneath of its belly. 
\vspace{-4mm}

\paragraph{Future work.} 
The next step involves combining semantic features with geometric ones, aiming for enhanced performance. Addressing potential noise in less visible areas of the distilled features is crucial, and we plan to explore the impact of refining features through geometric smoothness energies, such as local conformality or isometry. Overcoming challenges related to point clouds and non-manifold meshes is essential, given that many traditional geometry processing approaches assume manifold meshes. Additionally, there is an intriguing prospect of extending these descriptors to accommodate volumetric inputs like NeRFs or distance fields.

\paragraph{Acknowledgments.} 
We thank Romy Williamson for her comments and suggestions. NM was supported by Marie Skłodowska-Curie grant agreement No.~956585, gifts from Adobe, and UCL AI Centre.

{
    \small
    \bibliographystyle{ieeenat_fullname}
    \bibliography{reference}

\begin{thebibliography}{70}
\providecommand{\natexlab}[1]{#1}
\providecommand{\url}[1]{\texttt{#1}}
\expandafter\ifx\csname urlstyle\endcsname\relax
  \providecommand{\doi}[1]{doi: #1}\else
  \providecommand{\doi}{doi: \begingroup \urlstyle{rm}\Url}\fi

\bibitem[Abdelreheem et~al.(2023)Abdelreheem, Eldesokey, Ovsjanikov, and Wonka]{abdelreheem2023zero}
Ahmed Abdelreheem, Abdelrahman Eldesokey, Maks Ovsjanikov, and Peter Wonka.
\newblock Zero-shot 3d shape correspondence.
\newblock \emph{arXiv preprint arXiv:2306.03253}, 2023.

\bibitem[Aigerman et~al.(2014)Aigerman, Poranne, and Lipman]{aigerman2014lifted}
Noam Aigerman, Roi Poranne, and Yaron Lipman.
\newblock Lifted bijections for low distortion surface mappings.
\newblock \emph{ACM Transactions on Graphics (TOG)}, 33\penalty0 (4):\penalty0 1--12, 2014.

\bibitem[Aigerman et~al.(2015)Aigerman, Poranne, and Lipman]{aigerman2015seamless}
Noam Aigerman, Roi Poranne, and Yaron Lipman.
\newblock Seamless surface mappings.
\newblock \emph{ACM Transactions on Graphics (TOG)}, 34\penalty0 (4):\penalty0 1--13, 2015.

\bibitem[Aigerman et~al.(2022)Aigerman, Gupta, Kim, Chaudhuri, Saito, and Groueix]{aigerman2022neural}
Noam Aigerman, Kunal Gupta, Vladimir~G Kim, Siddhartha Chaudhuri, Jun Saito, and Thibault Groueix.
\newblock Neural jacobian fields: Learning intrinsic mappings of arbitrary meshes.
\newblock \emph{SIGGRAPH}, 2022.

\bibitem[Amir et~al.(2022)Amir, Gandelsman, Bagon, and Dekel]{amir2021deep}
Shir Amir, Yossi Gandelsman, Shai Bagon, and Tali Dekel.
\newblock Deep vit features as dense visual descriptors.
\newblock \emph{ECCVW What is Motion For?}, 2022.

\bibitem[Aubry et~al.(2011{\natexlab{a}})Aubry, Schlickewei, and Cremers]{wks}
Mathieu Aubry, Ulrich Schlickewei, and Daniel Cremers.
\newblock The wave kernel signature: A quantum mechanical approach to shape analysis.
\newblock In \emph{2011 IEEE international conference on computer vision workshops (ICCV workshops)}, pages 1626--1633. IEEE, 2011{\natexlab{a}}.

\bibitem[Aubry et~al.(2011{\natexlab{b}})Aubry, Schlickewei, and Cremers]{wks11}
Mathieu Aubry, Ulrich Schlickewei, and Daniel Cremers.
\newblock The wave kernel signature: A quantum mechanical approach to shape analysis.
\newblock In \emph{2011 IEEE International Conference on Computer Vision Workshops (ICCV Workshops)}, pages 1626--1633, 2011{\natexlab{b}}.

\bibitem[Bogo et~al.(2014)Bogo, Romero, Loper, and Black]{faust}
Federica Bogo, Javier Romero, Matthew Loper, and Michael~J Black.
\newblock Faust: Dataset and evaluation for 3d mesh registration.
\newblock In \emph{Proc. {CVPR}}, pages 3794--3801, 2014.

\bibitem[Bronstein et~al.(2008)Bronstein, Bronstein, and Kimmel]{tosca}
Alexander~M Bronstein, Michael~M Bronstein, and Ron Kimmel.
\newblock \emph{Numerical geometry of non-rigid shapes}.
\newblock Springer Science \& Business Media, 2008.

\bibitem[Cao and Bernard(2023)]{cao2023self}
Dongliang Cao and Florian Bernard.
\newblock Self-supervised learning for multimodal non-rigid 3d shape matching.
\newblock In \emph{Proceedings of the IEEE/CVF Conference on Computer Vision and Pattern Recognition}, pages 17735--17744, 2023.

\bibitem[Caron et~al.(2021)Caron, Touvron, Misra, J{\'e}gou, Mairal, Bojanowski, and Joulin]{dino}
Mathilde Caron, Hugo Touvron, Ishan Misra, Herv{\'e} J{\'e}gou, Julien Mairal, Piotr Bojanowski, and Armand Joulin.
\newblock Emerging properties in self-supervised vision transformers.
\newblock In \emph{Proceedings of the IEEE/CVF international conference on computer vision}, pages 9650--9660, 2021.

\bibitem[Chen et~al.(2023)Chen, Siddiqui, Lee, Tulyakov, and Nie{\ss}ner]{chen2023text2tex}
Dave~Zhenyu Chen, Yawar Siddiqui, Hsin-Ying Lee, Sergey Tulyakov, and Matthias Nie{\ss}ner.
\newblock Text2tex: Text-driven texture synthesis via diffusion models.
\newblock \emph{arXiv preprint arXiv:2303.11396}, 2023.

\bibitem[Decatur et~al.(2023)Decatur, Lang, and Hanocka]{decatur20233d}
Dale Decatur, Itai Lang, and Rana Hanocka.
\newblock 3d highlighter: Localizing regions on 3d shapes via text descriptions.
\newblock In \emph{Proceedings of the IEEE/CVF Conference on Computer Vision and Pattern Recognition}, pages 20930--20939, 2023.

\bibitem[Deng et~al.(2023)Deng, Wang, Lu, He, Zhang, Yu, and Zhang]{se-ornet}
Jiacheng Deng, Chuxin Wang, Jiahao Lu, Jianfeng He, Tianzhu Zhang, Jiyang Yu, and Zhe Zhang.
\newblock Se-ornet: Self-ensembling orientation-aware network for unsupervised point cloud shape correspondence.
\newblock In \emph{Proceedings of the IEEE/CVF Conference on Computer Vision and Pattern Recognition}, pages 5364--5373, 2023.

\bibitem[Deprelle et~al.(2019)Deprelle, Groueix, Fisher, Kim, Russell, and Aubry]{elementary}
Theo Deprelle, Thibault Groueix, Matthew Fisher, Vladimir Kim, Bryan Russell, and Mathieu Aubry.
\newblock Learning elementary structures for 3d shape generation and matching.
\newblock \emph{Advances in Neural Information Processing Systems}, 32, 2019.

\bibitem[Donati et~al.(2020)Donati, Sharma, and Ovsjanikov]{geomfmap}
Nicolas Donati, Abhishek Sharma, and Maks Ovsjanikov.
\newblock Deep geometric functional maps: Robust feature learning for shape correspondence.
\newblock In \emph{Proceedings of the IEEE/CVF Conference on Computer Vision and Pattern Recognition (CVPR)}, 2020.

\bibitem[Dyke et~al.(2020)Dyke, Lai, Rosin, Zappal{\`a}, Dykes, Guo, Li, Marin, Melzi, and Yang]{shrec20}
Roberto~M Dyke, Yu-Kun Lai, Paul~L Rosin, Stefano Zappal{\`a}, Seana Dykes, Daoliang Guo, Kun Li, Riccardo Marin, Simone Melzi, and Jingyu Yang.
\newblock Shrec’20: Shape correspondence with non-isometric deformations.
\newblock \emph{Computers \& Graphics}, 92:\penalty0 28--43, 2020.

\bibitem[Eisenberger et~al.(2019)Eisenberger, L{\"a}hner, and Cremers]{eisenberger2019divergence}
Marvin Eisenberger, Zorah L{\"a}hner, and Daniel Cremers.
\newblock Divergence-free shape correspondence by deformation.
\newblock In \emph{Computer Graphics Forum}, pages 1--12. Wiley Online Library, 2019.

\bibitem[Fischer et~al.(2024)Fischer, Li, Nguyen-Phuoc, Bozic, Dong, Marshall, and Ritschel]{fischer2024nerf}
Michael Fischer, Zhengqin Li, Thu Nguyen-Phuoc, Aljaz Bozic, Zhao Dong, Carl Marshall, and Tobias Ritschel.
\newblock Nerf analogies: Example-based visual attribute transfer for nerfs.
\newblock \emph{arXiv preprint arXiv:2402.08622}, 2024.

\bibitem[Giorgi et~al.(2007)Giorgi, Biasotti, and Paraboschi]{shrec2007}
Daniela Giorgi, Silvia Biasotti, and Laura Paraboschi.
\newblock Shape retrieval contest 2007: Watertight models track.
\newblock \emph{SHREC competition}, 8\penalty0 (7):\penalty0 7, 2007.

\bibitem[Groueix et~al.(2018{\natexlab{a}})Groueix, Fisher, Kim, Russell, and Aubry]{3d-coded}
Thibault Groueix, Matthew Fisher, Vladimir~G Kim, Bryan~C Russell, and Mathieu Aubry.
\newblock 3d-coded: 3d correspondences by deep deformation.
\newblock In \emph{Proceedings of the european conference on computer vision (ECCV)}, pages 230--246, 2018{\natexlab{a}}.

\bibitem[Groueix et~al.(2018{\natexlab{b}})Groueix, Fisher, Kim, Russell, and Aubry]{surreal}
Thibault Groueix, Matthew Fisher, Vladimir~G Kim, Bryan~C Russell, and Mathieu Aubry.
\newblock 3d-coded: 3d correspondences by deep deformation.
\newblock In \emph{Proceedings of the european conference on computer vision (ECCV)}, pages 230--246, 2018{\natexlab{b}}.

\bibitem[Hadsell et~al.(2006)Hadsell, Chopra, and LeCun]{hadsell2006dimensionality}
Raia Hadsell, Sumit Chopra, and Yann LeCun.
\newblock Dimensionality reduction by learning an invariant mapping.
\newblock In \emph{2006 IEEE computer society conference on computer vision and pattern recognition (CVPR'06)}, pages 1735--1742. IEEE, 2006.

\bibitem[Huang et~al.(2017)Huang, Kalogerakis, Chaudhuri, Ceylan, Kim, and Yumer]{10.1145/3137609}
Haibin Huang, Evangelos Kalogerakis, Siddhartha Chaudhuri, Duygu Ceylan, Vladimir~G. Kim, and Ersin Yumer.
\newblock Learning local shape descriptors from part correspondences with multiview convolutional networks.
\newblock \emph{ACM Trans. Graph.}, 37\penalty0 (1), 2017.

\bibitem[Kazhdan and Hoppe(2013)]{screened-poisson}
Michael Kazhdan and Hugues Hoppe.
\newblock Screened poisson surface reconstruction.
\newblock \emph{ACM Transactions on Graphics (ToG)}, 32\penalty0 (3):\penalty0 1--13, 2013.

\bibitem[Kirillov et~al.(2023)Kirillov, Mintun, Ravi, Mao, Rolland, Gustafson, Xiao, Whitehead, Berg, Lo, et~al.]{kirillov2023segment}
Alexander Kirillov, Eric Mintun, Nikhila Ravi, Hanzi Mao, Chloe Rolland, Laura Gustafson, Tete Xiao, Spencer Whitehead, Alexander~C Berg, Wan-Yen Lo, et~al.
\newblock Segment anything.
\newblock \emph{arXiv preprint arXiv:2304.02643}, 2023.

\bibitem[Kobayashi et~al.(2022)Kobayashi, Matsumoto, and Sitzmann]{kobayashi2022decomposing}
Sosuke Kobayashi, Eiichi Matsumoto, and Vincent Sitzmann.
\newblock Decomposing nerf for editing via feature field distillation.
\newblock \emph{Advances in Neural Information Processing Systems}, 35:\penalty0 23311--23330, 2022.

\bibitem[Komarichev et~al.(2019)Komarichev, Zhong, and Hua]{komarichev2019cnn}
Artem Komarichev, Zichun Zhong, and Jing Hua.
\newblock A-cnn: Annularly convolutional neural networks on point clouds.
\newblock In \emph{Proceedings of the IEEE/CVF conference on computer vision and pattern recognition}, pages 7421--7430, 2019.

\bibitem[Kraevoy and Sheffer(2004)]{kraevoy2004cross}
Vladislav Kraevoy and Alla Sheffer.
\newblock Cross-parameterization and compatible remeshing of 3d models.
\newblock \emph{ACM Transactions on Graphics (ToG)}, 23\penalty0 (3):\penalty0 861--869, 2004.

\bibitem[Krizhevsky et~al.(2012)Krizhevsky, Sutskever, and Hinton]{alexnet}
Alex Krizhevsky, Ilya Sutskever, and Geoffrey~E Hinton.
\newblock Imagenet classification with deep convolutional neural networks.
\newblock \emph{Advances in neural information processing systems}, 25, 2012.

\bibitem[Lang et~al.(2021)Lang, Ginzburg, Avidan, and Raviv]{dpc}
Itai Lang, Dvir Ginzburg, Shai Avidan, and Dan Raviv.
\newblock Dpc: Unsupervised deep point correspondence via cross and self construction.
\newblock In \emph{2021 International Conference on 3D Vision (3DV)}, pages 1442--1451. IEEE, 2021.

\bibitem[Lee et~al.(1999)Lee, Dobkin, Sweldens, and Schr{\"o}der]{lee1999multiresolution}
Aaron~WF Lee, David Dobkin, Wim Sweldens, and Peter Schr{\"o}der.
\newblock Multiresolution mesh morphing.
\newblock In \emph{Proceedings of the 26th annual conference on Computer graphics and interactive techniques}, pages 343--350, 1999.

\bibitem[L{\'e}vy et~al.(2023)L{\'e}vy, Petitjean, Ray, and Maillot]{levy2023least}
Bruno L{\'e}vy, Sylvain Petitjean, Nicolas Ray, and J{\'e}rome Maillot.
\newblock Least squares conformal maps for automatic texture atlas generation.
\newblock In \emph{Seminal Graphics Papers: Pushing the Boundaries, Volume 2}, pages 193--202. 2023.

\bibitem[Lin and Lee(2021)]{Lin_2021_CVPR}
Jiahao Lin and Gim~Hee Lee.
\newblock Multi-view multi-person 3d pose estimation with plane sweep stereo.
\newblock In \emph{Proceedings of the IEEE/CVF Conference on Computer Vision and Pattern Recognition (CVPR)}, pages 11886--11895, 2021.

\bibitem[Lipman(2012)]{lipman2012bounded}
Yaron Lipman.
\newblock Bounded distortion mapping spaces for triangular meshes.
\newblock \emph{ACM Transactions on Graphics (TOG)}, 31\penalty0 (4):\penalty0 1--13, 2012.

\bibitem[Litany et~al.(2017)Litany, Remez, Rodola, Bronstein, and Bronstein]{fmnet}
Or Litany, Tal Remez, Emanuele Rodola, Alex Bronstein, and Michael Bronstein.
\newblock Deep functional maps: Structured prediction for dense shape correspondence.
\newblock In \emph{Proceedings of the IEEE international conference on computer vision}, pages 5659--5667, 2017.

\bibitem[Luo et~al.(2023)Luo, Dunlap, Park, Holynski, and Darrell]{hyperfeatures}
Grace Luo, Lisa Dunlap, Dong~Huk Park, Aleksander Holynski, and Trevor Darrell.
\newblock Diffusion hyperfeatures: Searching through time and space for semantic correspondence.
\newblock \emph{arXiv preprint arXiv:2305.14334}, 2023.

\bibitem[Mandad et~al.(2017)Mandad, Cohen-Steiner, Kobbelt, Alliez, and Desbrun]{mandad2017variance}
Manish Mandad, David Cohen-Steiner, Leif Kobbelt, Pierre Alliez, and Mathieu Desbrun.
\newblock Variance-minimizing transport plans for inter-surface mapping.
\newblock \emph{ACM Transactions on Graphics (ToG)}, 36\penalty0 (4):\penalty0 1--14, 2017.

\bibitem[Melzi et~al.(2019)Melzi, Marin, Rodol{\`a}, Castellani, Ren, Poulenard, Wonka, and Ovsjanikov]{shrec'19}
Simone Melzi, Riccardo Marin, Emanuele Rodol{\`a}, Umberto Castellani, Jing Ren, Adrien Poulenard, Peter Wonka, and Maks Ovsjanikov.
\newblock Shrec 2019: Matching humans with different connectivity.
\newblock In \emph{Eurographics Workshop on 3D Object Retrieval}, page~3. The Eurographics Association, 2019.

\bibitem[Morreale et~al.(2021)Morreale, Aigerman, Kim, and Mitra]{morreale2021neural}
Luca Morreale, Noam Aigerman, Vladimir~G Kim, and Niloy~J Mitra.
\newblock Neural surface maps.
\newblock In \emph{Proceedings of the IEEE/CVF Conference on Computer Vision and Pattern Recognition}, pages 4639--4648, 2021.

\bibitem[Oquab et~al.(2023)Oquab, Darcet, Moutakanni, Vo, Szafraniec, Khalidov, Fernandez, Haziza, Massa, El-Nouby, Howes, Huang, Xu, Sharma, Li, Galuba, Rabbat, Assran, Ballas, Synnaeve, Misra, Jegou, Mairal, Labatut, Joulin, and Bojanowski]{dinov2}
Maxime Oquab, Timothée Darcet, Theo Moutakanni, Huy~V. Vo, Marc Szafraniec, Vasil Khalidov, Pierre Fernandez, Daniel Haziza, Francisco Massa, Alaaeldin El-Nouby, Russell Howes, Po-Yao Huang, Hu Xu, Vasu Sharma, Shang-Wen Li, Wojciech Galuba, Mike Rabbat, Mido Assran, Nicolas Ballas, Gabriel Synnaeve, Ishan Misra, Herve Jegou, Julien Mairal, Patrick Labatut, Armand Joulin, and Piotr Bojanowski.
\newblock Dinov2: Learning robust visual features without supervision, 2023.

\bibitem[Ovsjanikov et~al.(2012)Ovsjanikov, Ben-Chen, Solomon, Butscher, and Guibas]{functionalmaps}
Maks Ovsjanikov, Mirela Ben-Chen, Justin Solomon, Adrian Butscher, and Leonidas Guibas.
\newblock Functional maps: a flexible representation of maps between shapes.
\newblock \emph{ACM Transactions on Graphics (ToG)}, 31\penalty0 (4):\penalty0 1--11, 2012.

\bibitem[Phong(1975)]{phong}
Bui~Tuong Phong.
\newblock Illumination for computer generated pictures.
\newblock \emph{Commun. ACM}, 18\penalty0 (6):\penalty0 311–317, 1975.

\bibitem[Qi et~al.(2017)Qi, Yi, Su, and Guibas]{pointnet++}
Charles~Ruizhongtai Qi, Li Yi, Hao Su, and Leonidas~J Guibas.
\newblock Pointnet++: Deep hierarchical feature learning on point sets in a metric space.
\newblock \emph{Advances in neural information processing systems}, 30, 2017.

\bibitem[Qu{\'e}au et~al.(2017)Qu{\'e}au, M{\'e}lou, Durou, and Cremers]{queau2017dense}
Yvain Qu{\'e}au, Jean M{\'e}lou, Jean-Denis Durou, and Daniel Cremers.
\newblock Dense multi-view 3d-reconstruction without dense correspondences.
\newblock 2017.

\bibitem[Rabinovich et~al.(2017)Rabinovich, Poranne, Panozzo, and Sorkine-Hornung]{rabinovich2017scalable}
Michael Rabinovich, Roi Poranne, Daniele Panozzo, and Olga Sorkine-Hornung.
\newblock Scalable locally injective mappings.
\newblock \emph{ACM Transactions on Graphics (TOG)}, 36\penalty0 (4):\penalty0 1, 2017.

\bibitem[Radford et~al.(2021)Radford, Kim, Hallacy, Ramesh, Goh, Agarwal, Sastry, Askell, Mishkin, Clark, et~al.]{clip}
Alec Radford, Jong~Wook Kim, Chris Hallacy, Aditya Ramesh, Gabriel Goh, Sandhini Agarwal, Girish Sastry, Amanda Askell, Pamela Mishkin, Jack Clark, et~al.
\newblock Learning transferable visual models from natural language supervision.
\newblock In \emph{International conference on machine learning}, pages 8748--8763. PMLR, 2021.

\bibitem[Ravi et~al.(2020)Ravi, Reizenstein, Novotny, Gordon, Lo, Johnson, and Gkioxari]{pytorch3d}
Nikhila Ravi, Jeremy Reizenstein, David Novotny, Taylor Gordon, Wan-Yen Lo, Justin Johnson, and Georgia Gkioxari.
\newblock Accelerating 3d deep learning with pytorch3d.
\newblock \emph{arXiv preprint arXiv:2007.08501}, 2020.

\bibitem[Richardson et~al.(2023)Richardson, Metzer, Alaluf, Giryes, and Cohen-Or]{richardson2023texture}
Elad Richardson, Gal Metzer, Yuval Alaluf, Raja Giryes, and Daniel Cohen-Or.
\newblock Texture: Text-guided texturing of 3d shapes.
\newblock \emph{arXiv preprint arXiv:2302.01721}, 2023.

\bibitem[Rombach et~al.(2022)Rombach, Blattmann, Lorenz, Esser, and Ommer]{stable-diffusion}
Robin Rombach, Andreas Blattmann, Dominik Lorenz, Patrick Esser, and Bj{\"o}rn Ommer.
\newblock High-resolution image synthesis with latent diffusion models.
\newblock In \emph{Proceedings of the IEEE/CVF conference on computer vision and pattern recognition}, pages 10684--10695, 2022.

\bibitem[Roufosse et~al.(2019)Roufosse, Sharma, and Ovsjanikov]{surfmnet}
Jean-Michel Roufosse, Abhishek Sharma, and Maks Ovsjanikov.
\newblock Unsupervised deep learning for structured shape matching.
\newblock In \emph{Proceedings of the IEEE/CVF International Conference on Computer Vision}, pages 1617--1627, 2019.

\bibitem[Schreiner et~al.(2004)Schreiner, Asirvatham, Praun, and Hoppe]{schreiner2004inter}
John Schreiner, Arul Asirvatham, Emil Praun, and Hugues Hoppe.
\newblock Inter-surface mapping.
\newblock In \emph{ACM SIGGRAPH 2004 Papers}, pages 870--877. 2004.

\bibitem[Sharma et~al.(2022)Sharma, Yin, Maji, Kalogerakis, Litany, and Fidler]{sharma2022mvdecor}
Gopal Sharma, Kangxue Yin, Subhransu Maji, Evangelos Kalogerakis, Or Litany, and Sanja Fidler.
\newblock Mvdecor: Multi-view dense correspondence learning for fine-grained 3d segmentation.
\newblock In \emph{European Conference on Computer Vision}, pages 550--567. Springer, 2022.

\bibitem[Song et~al.(2020)Song, Meng, and Ermon]{ddim}
Jiaming Song, Chenlin Meng, and Stefano Ermon.
\newblock Denoising diffusion implicit models.
\newblock \emph{arXiv preprint arXiv:2010.02502}, 2020.

\bibitem[Sorkine and Alexa(2007)]{sorkine2007rigid}
Olga Sorkine and Marc Alexa.
\newblock As-rigid-as-possible surface modeling.
\newblock In \emph{Symposium on Geometry processing}, pages 109--116. Citeseer, 2007.

\bibitem[Su et~al.(2015)Su, Maji, Kalogerakis, and Learned-Miller]{mvcnn}
Hang Su, Subhransu Maji, Evangelos Kalogerakis, and Erik Learned-Miller.
\newblock Multi-view convolutional neural networks for 3d shape recognition.
\newblock In \emph{Proceedings of the IEEE International Conference on Computer Vision (ICCV)}, 2015.

\bibitem[Tang et~al.(2023)Tang, Jia, Wang, Phoo, and Hariharan]{dift}
Luming Tang, Menglin Jia, Qianqian Wang, Cheng~Perng Phoo, and Bharath Hariharan.
\newblock Emergent correspondence from image diffusion.
\newblock \emph{arXiv preprint arXiv:2306.03881}, 2023.

\bibitem[Tombari et~al.(2010)Tombari, Salti, and Di~Stefano]{tombari2010unique}
Federico Tombari, Samuele Salti, and Luigi Di~Stefano.
\newblock Unique signatures of histograms for local surface description.
\newblock In \emph{Computer Vision--ECCV 2010: 11th European Conference on Computer Vision, Heraklion, Crete, Greece, September 5-11, 2010, Proceedings, Part III 11}, pages 356--369. Springer, 2010.

\bibitem[Tran et~al.(2022)Tran, Hua, Tran, and Hoai]{tran2022self}
Bach Tran, Binh-Son Hua, Anh~Tuan Tran, and Minh Hoai.
\newblock Self-supervised learning with multi-view rendering for 3d point cloud analysis.
\newblock In \emph{Proceedings of the Asian Conference on Computer Vision}, pages 3086--3103, 2022.

\bibitem[Wang et~al.(2022)Wang, Cai, and Wang]{wang2022multi}
Wenju Wang, Yu Cai, and Tao Wang.
\newblock Multi-view dual attention network for 3d object recognition.
\newblock \emph{Neural Computing and Applications}, 34\penalty0 (4):\penalty0 3201--3212, 2022.

\bibitem[Wang and Solomon(2019)]{dcp}
Yue Wang and Justin~M Solomon.
\newblock Deep closest point: Learning representations for point cloud registration.
\newblock In \emph{Proceedings of the IEEE/CVF international conference on computer vision}, pages 3523,3526,3527, 2019.

\bibitem[Wang et~al.(2013)Wang, Gong, Wang, Cohen-Or, Zhang, and Chen]{projectiveAnalysis:13}
Yunhai Wang, Minglun Gong, Tianhua Wang, Daniel Cohen-Or, Hao Zhang, and Baoquan Chen.
\newblock Projective analysis for 3d shape segmentation.
\newblock \emph{ACM TOG}, 32\penalty0 (6), 2013.

\bibitem[Wang et~al.(2019)Wang, Sun, Liu, Sarma, Bronstein, and Solomon]{dgcnn}
Yue Wang, Yongbin Sun, Ziwei Liu, Sanjay~E Sarma, Michael~M Bronstein, and Justin~M Solomon.
\newblock Dynamic graph cnn for learning on point clouds.
\newblock \emph{ACM Transactions on Graphics (tog)}, 38\penalty0 (5):\penalty0 1--12, 2019.

\bibitem[Xu et~al.(2021)Xu, Zheng, Xu, Quan, and Ling]{9442303}
Yong Xu, Chaoda Zheng, Ruotao Xu, Yuhui Quan, and Haibin Ling.
\newblock Multi-view 3d shape recognition via correspondence-aware deep learning.
\newblock \emph{IEEE Transactions on Image Processing}, 30:\penalty0 5299--5312, 2021.

\bibitem[Yew and Lee(2020)]{rpmnet}
Zi~Jian Yew and Gim~Hee Lee.
\newblock Rpm-net: Robust point matching using learned features.
\newblock In \emph{Proceedings of the IEEE/CVF conference on computer vision and pattern recognition}, pages 11824,11827, 2020.

\bibitem[Yu et~al.(2020)Yu, Yang, Fan, and Wei]{yu2020latent}
Qian Yu, Chengzhuan Yang, Honghui Fan, and Hui Wei.
\newblock Latent-mvcnn: 3d shape recognition using multiple views from pre-defined or random viewpoints.
\newblock \emph{Neural Processing Letters}, 52:\penalty0 581--602, 2020.

\bibitem[Zeng et~al.(2021)Zeng, Qian, Zhu, Hou, Yuan, and He]{corrnet3d}
Yiming Zeng, Yue Qian, Zhiyu Zhu, Junhui Hou, Hui Yuan, and Ying He.
\newblock Corrnet3d: Unsupervised end-to-end learning of dense correspondence for 3d point clouds.
\newblock In \emph{Proceedings of the IEEE/CVF Conference on Computer Vision and Pattern Recognition}, pages 6052--6061, 2021.

\bibitem[Zhang et~al.(2023)Zhang, Herrmann, Hur, Cabrera, Jampani, Sun, and Yang]{zhang2023tale}
Junyi Zhang, Charles Herrmann, Junhwa Hur, Luisa~Polania Cabrera, Varun Jampani, Deqing Sun, and Ming-Hsuan Yang.
\newblock A tale of two features: Stable diffusion complements dino for zero-shot semantic correspondence.
\newblock \emph{arXiv preprint arXiv:2305.15347}, 2023.

\bibitem[Zhang and Agrawala(2023)]{controlnet}
Lvmin Zhang and Maneesh Agrawala.
\newblock Adding conditional control to text-to-image diffusion models.
\newblock \emph{arXiv preprint arXiv:2302.05543}, 2023.

\bibitem[Zuffi et~al.(2017)Zuffi, Kanazawa, Jacobs, and Black]{smal}
Silvia Zuffi, Angjoo Kanazawa, David~W Jacobs, and Michael~J Black.
\newblock 3d menagerie: Modeling the 3d shape and pose of animals.
\newblock In \emph{Proceedings of the IEEE conference on computer vision and pattern recognition}, pages 6365--6373, 2017.

\end{thebibliography}
}

\clearpage
\maketitlesupplementary

\section{Implementation Details}
\textbf{Rendering.} The multi-view rendering is implemented using PyTorch 3D~\cite{pytorch3d} with Hard Phong~\cite{phong} shading and appropriate lighting. We place the light source in the direction of the camera to ensure coherent lighting across views and set a bin size of 1 to capture enough details during rasterization. To render the shape from all angles, we vary the elevation ($\theta$) and azimuthal ($\phi$) angles in linearly spaced intervals between 0\degree-360\degree.\\
\textbf{Diffusion.} We use simple text prompts such as `human' for SHREC'19 or animal name for TOSCA and SHREC'20 (e.g.,  `dog', `cat', etc.) to guide the diffusion model to produce textured renderings. All text prompts are appended with `best quality, highly detailed, photorealistic' to serve as positive prompt. We further use the negative prompt, `lowres, low quality, monochrome' to prevent the diffusion model from producing poor quality textures.\\
\textbf{Point Clouds.} When applying our method on point clouds, \sname requires point clouds to have enough density to produce a (mostly) smooth and continuous depth map to accurately condition the diffusion model. For the SHREC'19 dataset, this would require about 8000 points. We replace normal maps ($\mathcal{D}$) in $G$ from \Cref{eq:coloreq} with Canny edge maps ($\mathcal{E}$) since accurate estimation of normals from a point cloud can be challenging. We compute the edge map from the depth map instead of the point cloud render because the depth map is smooth and hence less noisy.\begin{equation} \label{eq:pc}
    G := \{\mathcal{E}(I^S_j),\mathcal{D}(I^S_j)\},
\end{equation}
\textbf{Code.} Our implementation can be found at \href{https://github.com/niladridutt/Diffusion-3D-Features}{https://github.com/niladridutt/Diffusion-3D-Features}
\textbf{Evaluation.} To have a shared baseline and maintain parity with previous works \cite{dpc,se-ornet,corrnet3d}, we downsample the original point cloud to 1024 points by random sampling. For our method, we use the complete mesh for rendering and compute descriptors for only 1024 points during the unprojection process. 

\section{Semantic Correspondence}
To showcase the semantic nature of \sname descriptors, we present heatmap visualizations in our \href{https://diff3f.github.io/}{project webpage}. For a query point in the source, we see semantically similar regions being highlighted in the target shape. For example, in the teddy $\rightarrow$ table pair, we see the legs of the teddy correspond with the legs of the table. We additionally showcase correspondence results on highly non-isometric pairs such as octopus $\rightarrow$ ant. We see that the head of the octopus maps to the head and tail of the ant, whereas the three legs of the octopus map to the two antennas and six legs of the ant.

\section{Evaluation on SHREC'07} 
We evaluate \sname and baseline methods on the SHREC Watertight 2007 \cite{shrec2007} dataset to show performance on diverse shape classes outside commonly tested human and animal shapes. The dataset comprises of 400 3D shapes of 20 classes spanning glasses, fish, plane, plier, etc.~with an average of 15 annotated correspondence points for each class. We skip all non human and four legged animals as these classes have been already evaluated (SHREC'19, TOSCA, SHREC'20). We additionally do not evaluate on spring and vase as these shapes lack annotations. The resulting dataset comprises of 320 shapes and we pair shapes from the same class to form 3040 test pairs, considerably larger than previous evaluations. Similar to previous evaluations, we choose DPC~\cite{dpc} and SE-ORNet~\cite{se-ornet} models trained on the SURREAL~\cite{surreal} dataset as it contains the largest number of training shapes. The weak performance of DPC in \Cref{table:shrec-07} highlights the importance of a general purpose correspondence method. We were unable to evaluate SE-ORNet on SHREC'07 as the method failed on shape classes with a low number of annotated points. 

\begin{table}[h]
\caption{\textbf{Evaluation on SHREC'07}. \sname (ours) attains 3.5x lower error compared to DPC. This highlights the importance of a general purpose correspondence method to adeptly work on a wide range of classes.}
\centering
\resizebox{\linewidth}{!}{
\begin{tabularx}{\linewidth} {c|c|c}
\footnotesize 
    \textbf{Method} & $acc \uparrow$ & $err \downarrow$ \\
    \midrule  \midrule
    DPC~\cite{dpc} & 35.63 & 4.91\\
    \hline
    SE-ORNet~\cite{se-ornet} & \xmark & \xmark \\
    \hline
    \sname (ours) & \textbf{52.73} & \textbf{1.41} \\
    \bottomrule
\end{tabularx}}
    \label{table:shrec-07}
\end{table}

\subsection{Robustness to Rotation} \label{sec:rotation}
We randomly rotate shapes in SHREC’19 and TOSCA datasets, sampling X and Y direction values from a uniform distribution in $U(-180\degree, 180\degree]$. Our method's performance is compared against DPC \cite{dpc} and SE-ORNet \cite{se-ornet} in \Cref{table:rotation}. SE-ORNet aligns shapes before computing correspondences for robustness, but this is ineffective for large rotations on SHREC’19 and TOSCA. In contrast, our method proves highly robust to substantial rotations, maintaining accuracy due to rotation invariance in the multi-view rendering and unprojection process, unlike baseline methods.

\begin{table}[h]
\vspace*{0.2in}
\caption{\textbf{Robustness to rotation.} Our method being based on multi-view rendering is invariant to rotation and sees no degradation in performance.}
\centering
\resizebox{\columnwidth}{!}{
\begin{tabularx}{\linewidth} {c|c|c|c|c}
\footnotesize 
    \textbf{Dataset} $\rightarrow$ & \multicolumn{2}{c|}{SHREC'19} & \multicolumn{2}{c}{TOSCA}  \\
    $\downarrow$ \textbf{Method} & $acc \uparrow$ & $err \downarrow$ & $acc \uparrow$ & $err \downarrow$ \\
    \midrule  \midrule
    DPC~\cite{dpc} & 8.83 & 9.03 & 3.89 & 17.87 \\
    \hline
    SE-ORNet~\cite{se-ornet} & 8.77 & 7.86 & 4.23 & 28.23 \\
    \hline
    \sname (ours) & \textbf{26.41} & \textbf{1.69} & \textbf{20.27} & \textbf{5.69} \\
    \bottomrule
\end{tabularx}}
    \label{table:rotation}
\end{table}

\section{Robustness to Text Prompt}
We evaluated our method on SHREC'19~\cite{shrec'19} using the prompt `human' against manually labeled `man' or `woman.' We do not see a significant change in performance, indicating our method is robust to prompt variations. We attribute this to additional channels like normal maps and depth maps that give hints about the object. We do not see a significant change in performance when using varied text prompts for different renderings either.

\section{Compute Time}
\sname distills 2D features from StableDiffusion~\cite{stable-diffusion} to 3D and hence does not require any further training. Our method takes about 2-3 minutes to compute descriptors of a shape for number of views ($n$) = 100 on a single Nvidia RTX 4090 GPU. The descriptors only need to be computed once and can be used without further optimization.

\end{document}